\documentclass{article}
\PassOptionsToPackage{numbers,compress}{natbib}
\usepackage[main, final]{neurips_2026}

\usepackage[utf8]{inputenc}
\usepackage[T1]{fontenc}
\usepackage{hyperref}
\usepackage{url}
\usepackage{booktabs}
\usepackage{amsfonts}
\usepackage{nicefrac}
\usepackage{microtype}
\usepackage{xcolor}
\usepackage{amsmath}
\usepackage{graphicx}
\usepackage{wrapfig}
\usepackage{array}
\usepackage{tabularx}
\usepackage{mathtools}

\title{Routers Learn the Geometry of Their Experts: \\
Geometric Coupling in Sparse Mixture-of-Experts}

\author{
    Sagi Ahrac\textsuperscript{*} \qquad
    Noya Hochwald\textsuperscript{*} \qquad
    Mor Geva \vspace{5pt} \\
    Blavatnik School of Computer Science and AI, Tel Aviv University \vspace{5pt} \\
    \texttt{\{sagiahrac@mail, noyahochwald@mail, morgeva@tauex\}.tau.ac.il}
}

\begin{document}

\maketitle

\begin{abstract}
Sparse Mixture-of-Experts (SMoE) models enable scaling language models efficiently, but training them remains challenging, as routing can collapse onto few experts and auxiliary load-balancing losses can reduce specialization. Motivated by these hurdles, we study how routing decisions in SMoEs are formed mechanistically. First, we reveal a geometric coupling between routers and their corresponding experts. For a given token, the router weights for the selected expert and the expert weights processing it receive gradients along the same input direction, differing only in scalar coefficients. Thus, matched router--expert directions accumulate the same routed token history. This theoretical coupling also appears empirically in routing dynamics. In a $1$B SMoE trained from scratch, higher router scores predict stronger expert neuron preactivations, showing that routing decisions are mirrored inside the selected expert. Next, we analyze the effects of auxiliary load balancing on the router--expert geometric coupling, showing that such losses break this structure by spreading input-directed gradients across router weights, making distinct router directions nearly three times more similar to each other. Last, we demonstrate the centrality of geometric coupling for effective routing with a parameter-free online K-Means router, in which each expert maintains a running average of the hidden states routed to it and tokens are assigned based on cosine similarity. Compared with auxiliary-loss and loss-free balancing, this router achieves the lowest load imbalance with only a modest perplexity increase, indicating that geometric coupling captures a substantial part of what the router learns. Overall, our results explain how routers form assignment geometry that supports an effective division of labor.
\end{abstract}

\section{Introduction}

Sparse Mixture-of-Experts (SMoE) has emerged as a leading architecture for scaling language model parameters without a proportional increase in inference latency \citep{shazeer2017outrageouslylargeneuralnetworks, fedus2022switchtransformersscalingtrillion, lepikhin2020gshardscalinggiantmodels}.
Recent implementations, such as DeepSeek-V3 \citep{deepseekai2025deepseekv3technicalreport} and OLMoE \citep{muennighoff2025olmoeopenmixtureofexpertslanguage}, match or outperform dense models while activating only a fraction of their total parameters.
These efficiency gains stem from a division of labor in which a \textit{router}, typically implemented as a small gating network, directs each input to a subset of independent \textit{expert} networks \citep{shazeer2017outrageouslylargeneuralnetworks, fedus2022switchtransformersscalingtrillion}.
Still, despite their wide adoption, training SMoEs with effective routing remains a challenge.
Without intervention, routing concentrates on a shrinking subset of experts, leading to representation collapse \citep{chi2022representationcollapsesparsemixture}.
Approaches to mitigate this apply auxiliary load-balancing losses
\citep{shazeer2017outrageouslylargeneuralnetworks,lepikhin2020gshardscalinggiantmodels,fedus2022switchtransformersscalingtrillion}, which encourage balanced routing but often reduce expert specialization \citep{wang2024auxiliarylossfreeloadbalancing,deepseekai2025deepseekv3technicalreport}.
Motivated by these pathologies, we seek to understand the inner dynamics of routing in SMoEs, focusing on how routing decisions are derived mechanistically.

We tackle this question from a geometric view. 
First, we analyze the gradients that shape both sides of the routing decision. Although routers and experts are often treated as separate modules~\citep{shazeer2017outrageouslylargeneuralnetworks,dai2022stablemoestableroutingstrategy}, SMoE gradients reveal a shared input-directed structure. For each assigned token, the router weights associated with the selected expert and the expert weights that process the token receive updates along the same input direction, differing only in scalar coefficients.
Notably, this alignment is not a generic consequence of joint training: a shared computation graph derives both modules to receive gradients, but does not imply that their updates accumulate along shared input directions. In SMoE layers, however, the chain rule enforces this proportional form, inducing a \emph{geometric coupling} where matched router--expert directions evolve as coupled accumulators of their shared routed token history.

Next, we investigate if this theoretical coupling translates into empirical routing dynamics.
To this end, we compare the router's score for an expert and that expert's neuron preactivations in response to the same token.
In a $1$B SMoE trained from scratch for approximately $50$B tokens, we find that experts ranked higher by the router consistently exhibit stronger preactivations than experts not selected by the router.
This shows that routing decisions are not merely an external assignment, but are mirrored in the computation of the selected expert, thus providing empirical evidence of the geometric coupling between the router and experts.

Having established the router--expert geometric coupling, we revisit common routing instabilities and load-balancing side effects.
Specifically, we study how the router's geometry is influenced by the common auxiliary load-balancing loss employed in existing models~\citep{shazeer2017outrageouslylargeneuralnetworks,lepikhin2020gshardscalinggiantmodels,fedus2022switchtransformersscalingtrillion}. This loss encourages balanced routing by penalizing uneven expert load.
From a theoretical point of view, this optimization sends input-directed gradients to every router weight vector on every token, regardless of which experts were chosen.
This is expected to unify different router directions over training, weakening the geometric coupling between the router and the experts.
We evaluate this prediction by training two $1$B SMoEs that differ only in the balancing rule, finding that distinct router weight vectors become nearly three times more similar with the auxiliary loss than without it.
These results show that auxiliary balancing regulates expert usage, yet breaks the geometric coupling, unifying expert-specific directions and eroding the specialization that coupling produces.

\begin{figure*}[t]
  \centering
  \vspace{0.4em}
    \includegraphics[width=0.9\textwidth]{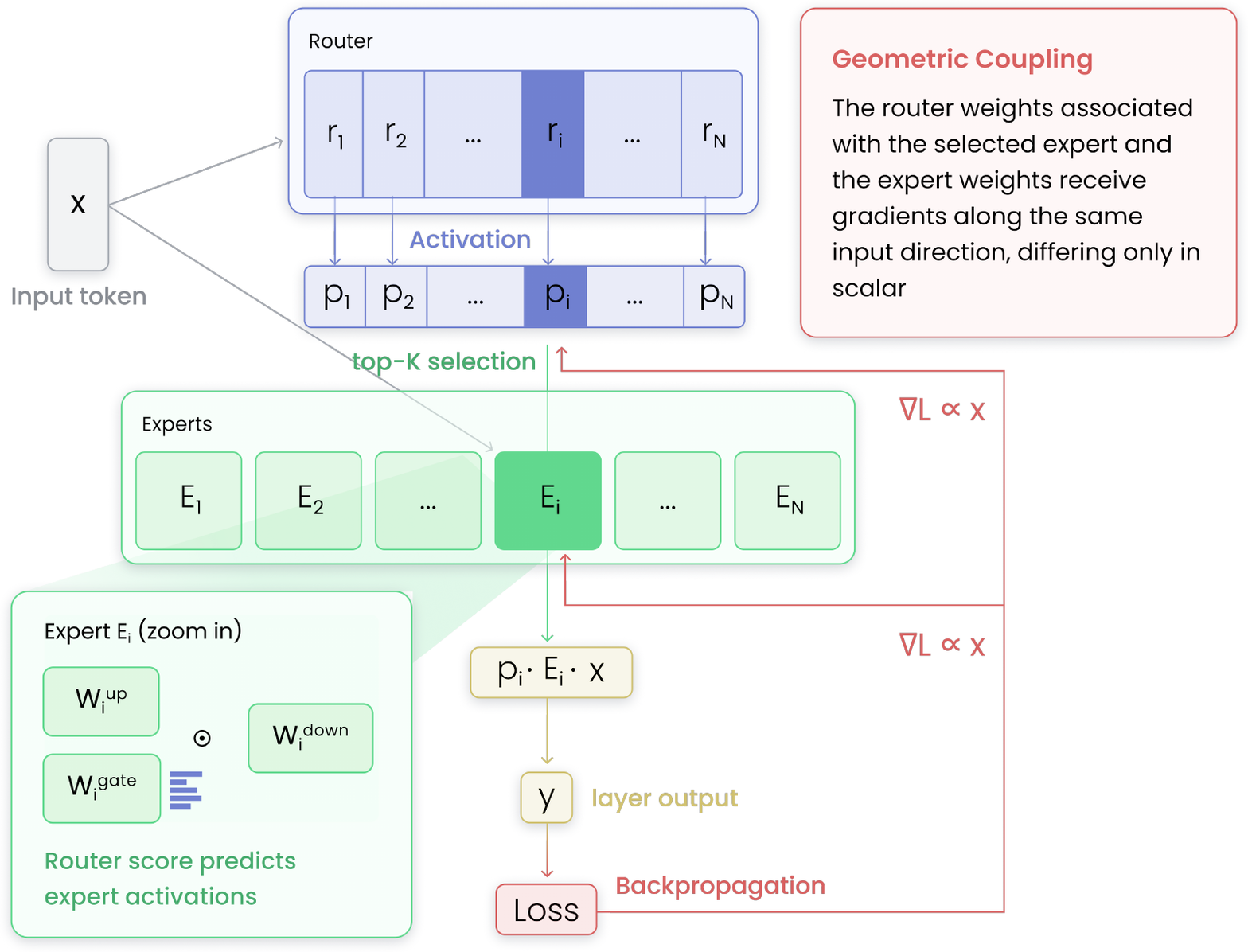}
    \caption{\textbf{Router--expert geometric coupling in SMoEs.}
The router scores a hidden state $\mathbf{x}$ and selects a sparse set of top-$K$ experts.
For each selected expert, the matched router direction and expert input-side weights receive backpropagation updates proportional to the same hidden-state direction $\mathbf{x}$.
Repeated updates make matched router--expert pairs accumulate a common routed-token history, which is read out at inference as higher router scores predicting stronger gate-neuron preactivations.}
  \label{fig:router_mechanics}
\end{figure*}

Last, we demonstrate the centrality of geometric coupling in routing by evaluating a parameter-free centroid router that follows this natural coupling.
Concretely, if router weights learn to summarize the hidden states assigned to each expert, then explicit summaries may recover much of their role.
Following this idea, we implement the centroid router using online K-Means~\citep{macqueen1967some}.
Each expert maintains a centroid of its assigned hidden states, and tokens are routed by cosine similarity to these centroids. On our $1$B SMoE setup, this router achieves the lowest load imbalance among all variants while maintaining comparable perplexity to the learned loss-free router. This indicates that geometric coupling is not only a byproduct of training, but a substantial component of what routers learn.

In conclusion, our work makes the following contributions:
\begin{enumerate}
    \item 
    We reveal a \emph{geometric coupling} between routers and experts in SMoEs, where matched router--expert pairs receive gradient updates along the same routed-token directions and evolve as coupled accumulators of their shared token history.
    \item 
    We show empirically that router scores are reflected in the selected expert's internal computation, with higher-ranked experts exhibiting stronger preactivations for the same tokens.
    \item 
    We show that common auxiliary load-balancing losses act directly on router geometry, injecting input-directed gradients into every router weight vector regardless of selection and making distinct router directions nearly three times more similar.
    \item 
    Motivated by coupling, we instantiate a simple online K-Means router in which non-learnable EMA centroids replace learned router weights, and tokens are routed by cosine similarity with sign-updated biases. This router achieves the lowest load imbalance among our variants with only a modest perplexity increase.
\end{enumerate}
Taken together, our results answer how routers form assignment geometry that supports an effective division of labor. More broadly, they suggest that future routing methods may benefit from preserving the natural router--expert geometry that emerges during training.
We release our code at \url{https://github.com/sagearc/router-expert-geometry}.

\section{SMoE architecture and training}
\label{sec:smoe_arch_and_training}

\paragraph{The SMoE architecture}

SMoEs are typically built on top of the Transformer architecture~\citep{vaswani2017attentionisallyouneed}, replacing feed-forward blocks with a set of $N$ distinct expert networks $\{E_1,\dots,E_N\}$ and a router $R$ that selects a small subset of experts for each input~\citep{shazeer2017outrageouslylargeneuralnetworks,lepikhin2020gshardscalinggiantmodels,fedus2022switchtransformersscalingtrillion,dai2024deepseekmoeultimateexpertspecialization}.
For a hidden state $\mathbf{x} \in \mathbb{R}^d$, the router computes one score per expert:
\begin{equation}
    \mathbf{p} = \sigma(\mathbf{z} + \mathbf{m}), \qquad \mathbf{z} = W_r\mathbf{x}.
\end{equation}
where $W_r \in \mathbb{R}^{N\times d}$ are the router weights, with the $i$-th row $\mathbf{r}_i$ corresponding to the expert $i$, $\mathbf{m}\in \mathbb{R}^{N}$ is a mask for selecting the top-$K$ entries in $\mathbf{z}$, and $\sigma$ is a nonlinear activation function.
Throughout the paper, we refer to $z_i$ and $p_i$ as the router's score and the routing weight of expert $i$, respectively, and denote the set of top-$K$ selected experts by $\mathcal{T}_K$.
The SMoE layer combines only the selected expert outputs:
\begin{equation}
\label{eq:smoe_output}
    y = \sum_{i\in\mathcal{T}_K} p_iE_i(\mathbf{x}).
\end{equation}

Each expert $E_i$ maps the hidden state through an intermediate expert dimension.
In the gated SwiGLU experts used by recent SMoEs~\citep{jiang2024mixtralexperts,muennighoff2025olmoeopenmixtureofexpertslanguage,deepseekai2025deepseekv3technicalreport}, this computation can be written as
\begin{equation}
    E_i(\mathbf{x}) =
    W^{\mathrm{down}}_{i}\!\left(
    \sigma(W^{\mathrm{gate}}_{i}\mathbf{x}) \odot W^{\mathrm{up}}_{i}\mathbf{x}
    \right),
\end{equation}
where $\sigma$ is the SiLU activation and $W^{\mathrm{gate}}_{i}, W^{\mathrm{up}}_{i} \in \mathbb{R}^{d_{\mathrm{i}}\times d}$ are input-side expert matrices with an intermediate dimension $d_{i}$. We refer to the vectors $W^{\mathrm{gate}}_{i}\mathbf{x}$ and $W^{\mathrm{up}}_{i}\mathbf{x}$ as the preactivations of the expert's gate- and up-neurons.

\paragraph{SMoE training and load balancing}
Training SMoEs requires both meaningful token-to-expert assignments and balanced expert utilization.
For a batch of training examples $\mathcal{B}$, let $f_i$ denote the fraction of top-$K$ assignments received by expert $i$:
\begin{equation}
    f_i =
    \frac{1}{K|\mathcal{B}|}
    \sum_{\mathbf{x} \in \mathcal{B}}
\mathbf{1}\{i \in \mathcal{T}_K\},
    \qquad
    \tau = \frac{1}{N}.
\end{equation}
Here $\mathcal{T}_K$ is the set of top-$K$ expert indices for $\mathbf{x}$, and $\tau$ is the target uniform utilization. Standard SMoE training often adds an auxiliary load-balancing loss~\citep{shazeer2017outrageouslylargeneuralnetworks,lepikhin2020gshardscalinggiantmodels,fedus2022switchtransformersscalingtrillion},
\begin{equation}
    \mathcal{L}
    =
    \mathcal{L}_{\mathrm{LM}}
    +
    \lambda_{\mathrm{aux}}\mathcal{L}_{\mathrm{balance}},
\end{equation}
where $\mathcal{L}_{\mathrm{balance}}$ penalizes the unequal distribution of tokens across experts, and $\lambda_{\mathrm{aux}}$ dictates the strength of this penalty. In contrast, recent loss-free methods instead add adaptive per-expert routing biases~\citep{wang2024auxiliarylossfreeloadbalancing,deepseekai2025deepseekv3technicalreport},
\begin{equation}
    \tilde{z}_i = z_i + b_i,
    \qquad
    b_i \leftarrow b_i + \gamma\,\mathrm{sign}(\tau - f_i),
\end{equation}
where $\gamma > 0$ is the bias update rate.
In our experiments, this bias affects top-$K$ selection but is not added to the expert weights used in Eq.~\eqref{eq:smoe_output}.

\section{Router--expert geometric coupling}\label{sec:co_evolution_hyp}

We formalize the gradient structure underlying the learned routing, showing that for a token representation $\mathbf{x}$ routed to expert $i$, both the corresponding router weight vector $\mathbf{r}_i$ and the input-side weights of expert $i$ receive updates proportional to $\mathbf{x}$. 
Over training, matched router--expert pairs accumulate the same routed token history with different weights, a structure that follows from the chain rule rather than from an added constraint.

To derive the coupling between router weight vectors and expert input matrices, we consider a standard backpropagation step through the SMoE layer in Eq.~\eqref{eq:smoe_output}.
On the expert side, each row of $W^{\mathrm{gate}}_{i}$ takes an inner product with $\mathbf{x}$ to produce one coordinate of the gate activations. Let $\mathbf{w}_{i,k}$ be $k$-th row of $W^{\mathrm{gate}}_{i}$, its respective gradient can be written as
\begin{equation}
    \nabla_{\mathbf{w}_{i,k}} \mathcal{L} = \delta_{i,k}\,\mathbf{x}^{\top} \propto \mathbf{x}^{\top},
\end{equation}
where $\delta_{i,k}$ collects all scalar factors other than the input direction. Therefore, the rows of 
$W^{\mathrm{gate}}_{i}$ evolve into weighted sums of the hidden states processed by expert $i$.
Notably, the same input-directed form also applies to $W^{\mathrm{up}}_{i}$, with a different scalar coefficient.
For a detailed mathematical derivation and an extension for models trained with AdamW updates, see Appendix~\ref{sec:appendix_derivation}.

Simultaneously, the router weight vector $\mathbf{r}_i$ follows the same input-directed update structure.
Given the logit $z_i = \mathbf{x}^{\top}\mathbf{r}_i$, the gradient with respect to $\mathbf{r}_i$ is
\begin{equation}
    \nabla_{\mathbf{r}_i} \mathcal{L} = \left( \frac{\partial \mathcal{L}}{\partial p_i} \frac{\partial p_i}{\partial z_i} \right) \mathbf{x} = \gamma_i \mathbf{x} \propto \mathbf{x},
\end{equation}
where $\gamma_i$ is the scalar coefficient for the router 
update (see Appendix~\ref{sec:router_appendix_derivation} for the full router-gradient derivation).  
For both the router and the expert input-side matrices, weights are updated by adding the hidden-state direction $\mathbf{x}$ when it reduces the loss and subtracting $\mathbf{x}$ when it increases it.
Thus, over training, matched router--expert pairs accumulate the same routed token history with different weights, making this coupling a direct consequence of backpropagation through the SMoE layer rather than an added constraint.
This shared gradient structure suggests that the paired router and expert weights should develop \textbf{aligned geometry in hidden space} (Figure~\ref{fig:router_mechanics}).

This accumulation is expert-specific, as experts outside the top-$K$ set do not contribute to the layer's output. For such an expert $j \notin \mathcal{T}_K$, the routing weight $p_j$ is zero, and its router weight vector $\mathbf{r}_j$ receives no gradient from this token. Each router weight vector therefore accumulates only the tokens routed to its expert, yielding the expert-specific accumulation that drives geometric coupling.
This view implies that router preference should be reflected inside the selected expert. Specifically, if $\mathbf{r}_i$ and the rows of $W^{\mathrm{gate}}_{i}$ are shaped by the same routed hidden states, then a token that receives a high router score for expert $i$ should also activate that expert's gate neurons more strongly. Section~\ref{sec:empirical_coupling} tests this relationship directly.

\section{Empirical evidence of geometric coupling}
\label{sec:empirical_coupling}

\begin{figure}[t]
  \centering
  \includegraphics[width=\linewidth]{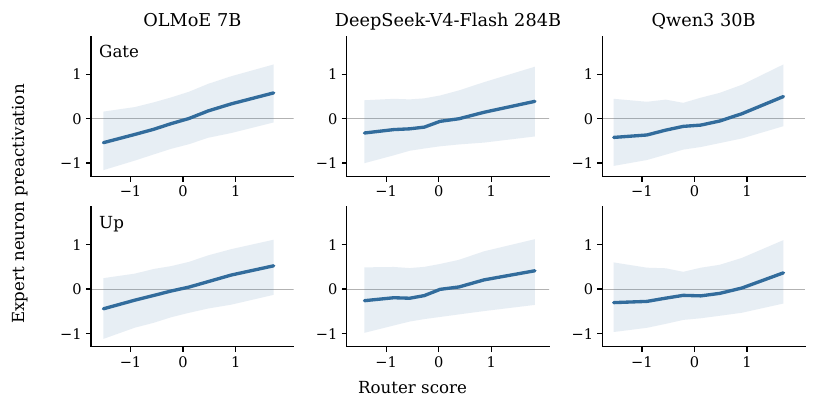}
  \caption{\textbf{Expert neuron preactivations increase with router score.} For each routed token--expert pair in OLMoE, DeepSeek, and Qwen, we compare the router score $z_i$ with $\|W^{\mathrm{gate}}_{i}\mathbf{x}\|_2$ (top) and $\|W^{\mathrm{up}}_{i}\mathbf{x}\|_2$ (bottom). For each model, scores and norms are normalized separately for each layer and expert before pooling. We observe that router scores and expert responses are correlated. Lines show median responses in eight score-quantile bins; shading shows the interquartile range.}
  \label{fig:expert_router_correlation}
\end{figure}

The derivation in Section~\ref{sec:co_evolution_hyp} shows that router scores and expert activations share a common geometry because both are shaped by the same routed hidden states. Here, we examine this relationship empirically, asking whether router preference is reflected inside the selected expert's computation.

For a token $\mathbf{x}$, the router score $z_i=\mathbf{r}_i^\top\mathbf{x}$ measures the router's preference for expert $i$.
For the same token, the expert's gate- and up-neuron preactivations are the coordinates of $\|W^{\mathrm{gate}}_{i}\mathbf{x}\|_2$ and $\|W^{\mathrm{up}}_{i}\mathbf{x}\|_2$, respectively. Importantly, these quantities are computed independently in the same forward pass, i.e., the router score is not used to compute the expert activations.
If $\mathbf{r}_i$ and the expert input-side vectors are shaped by the same routed inputs during training, then higher router scores should correspond to stronger expert preactivations.

We evaluate OLMoE \citep{muennighoff2025olmoeopenmixtureofexpertslanguage}, DeepSeek-V4-Flash \citep{deepseekai2026deepseekv4highlyefficientmilliontoken}, and Qwen3-30B-A3B-Base \citep{yang2025qwen3technicalreport} on the final token of each of the same 1,000 English Wikipedia titles. For each selected token--expert pair, we record the router score and the $L_2$ norms of the gate and up preactivations. For each model, scores and norms are standardized separately within each layer and expert before pooling. Further details on the measurement procedure are provided in Appendix~\ref{app:preactivation}.

\paragraph{Router preferences predict expert responses.}
Figure~\ref{fig:expert_router_correlation} shows that in all three models, higher router scores are associated with stronger expert responses in both the gate and up projections, as predicted by the router--expert coupling.
This is consistent with the shared training history of $\mathbf{r}_i$ and the expert’s input vectors, which are shaped by the same routed inputs. Namely, inputs that shaped $\mathbf{r}_i$ during training also shaped the input vectors of $W^{\mathrm{gate}}_{i}$ and $W^{\mathrm{up}}_{i}$, so a new token aligned with that history receives a higher router score and produces stronger expert preactivations.
Notably, these findings complement the correlation between router and gate-projection weights observed by \citet{lo-etal-2025-closer} by showing that router--expert correlation is not only visible in trained weights, but also appears in token-level expert activations during routing.

In Appendix~\ref{app:dynamics}, we further examine how this relationship develops during training. In OLMoE, the correlation is already present at the first measured checkpoint, after 5,000 training steps. We also track this in a model we trained from initialization, and observe that this correlation rises early in training and then partially declines. Notably, the correlation is higher for routed pairs than for non-routed or randomly paired router--expert pairs.

\paragraph{Router-aligned information causally influences predictions.}
Geometric coupling suggests that the router direction may capture information used by the expert itself, beyond selecting the tokens it receives. We examine this contribution while holding expert selection and mixture weights fixed, removing the router-aligned component from selected experts' inputs at one layer at a time. For comparison, we remove components along 32 orthogonal directions derived from the same experts' gate and up weights. In 15 of 16 OLMoE layers, router-direction removal produces a loss increase that is 1.6--54$\times$ higher than the median increase under these controls (see more details in Appendix~\ref{app:causal}).

\section{Auxiliary load-balancing breaks geometric coupling}
\label{sec:aux_loss_collapse}

The previous section showed that router preferences are reflected inside experts when coupling is preserved. We now examine what happens to the geometric coupling under the auxiliary load-balancing loss, a widely used practice for avoiding load-imbalance across experts.

\paragraph{The auxiliary loss breaks expert-specific accumulation}

In Section~\ref{sec:co_evolution_hyp} we showed that, under the language-modeling loss, only selected experts contribute to the SMoE layer output. For an unselected expert $j \notin \mathcal{T}_K$, the routing weight $p_j$ is zero and $\mathbf{r}_j$ receives no gradients from this token. This preserves expert-specific accumulation.
Conversely, we observe that the standard auxiliary load-balancing loss~\citep{shazeer2017outrageouslylargeneuralnetworks,fedus2022switchtransformersscalingtrillion} breaks this expert-specific structure:
\begin{equation}
\mathcal{L}_{\mathrm{balance}} = N \sum_{i=1}^{N} f_i\, P_i, \qquad
P_i = \frac{1}{|\mathcal{B}|}\sum_{\mathbf{x} \in \mathcal{B}} p_i,
\end{equation}
where $P_i$ is computed from an \emph{unmasked} softmax over all $N$ experts, i.e., $\text{softmax}(\mathbf{z})$, and therefore depends on every router weight vector, not only on the selected experts. The chain rule yields, for every token $\mathbf{x}$ in the batch and every expert $j$,
\begin{equation}
\nabla_{\mathbf{r}_j}\mathcal{L}_{\mathrm{balance}} = \beta_j\, \mathbf{x},
\qquad
\beta_j \neq 0
\end{equation}
where $\beta_j$ is a scalar coefficient that depends on the auxiliary loss and routing probabilities.
Every router weight vector therefore absorbs an input-directed contribution from every token, regardless of which experts were chosen. Instead of accumulating only the tokens routed to expert $j$, $\mathbf{r}_j$ also accumulates signal from the global token stream.
Notably, these are the ``interference gradients'' of \citet{wang2024auxiliarylossfreeloadbalancing}. Since hidden states at a given layer are not centered across the token stream, their average over many tokens can be nonzero. Because the auxiliary-loss gradient on each router weight vector is proportional to $\mathbf{x}$ on every token, all router weight vectors can accumulate a shared mean component, and across training this shared component can pull the vectors together, giving a coupling-failure signature which we now measure directly.

\begin{figure}[t]
  \centering
  \includegraphics[width=0.9\columnwidth]{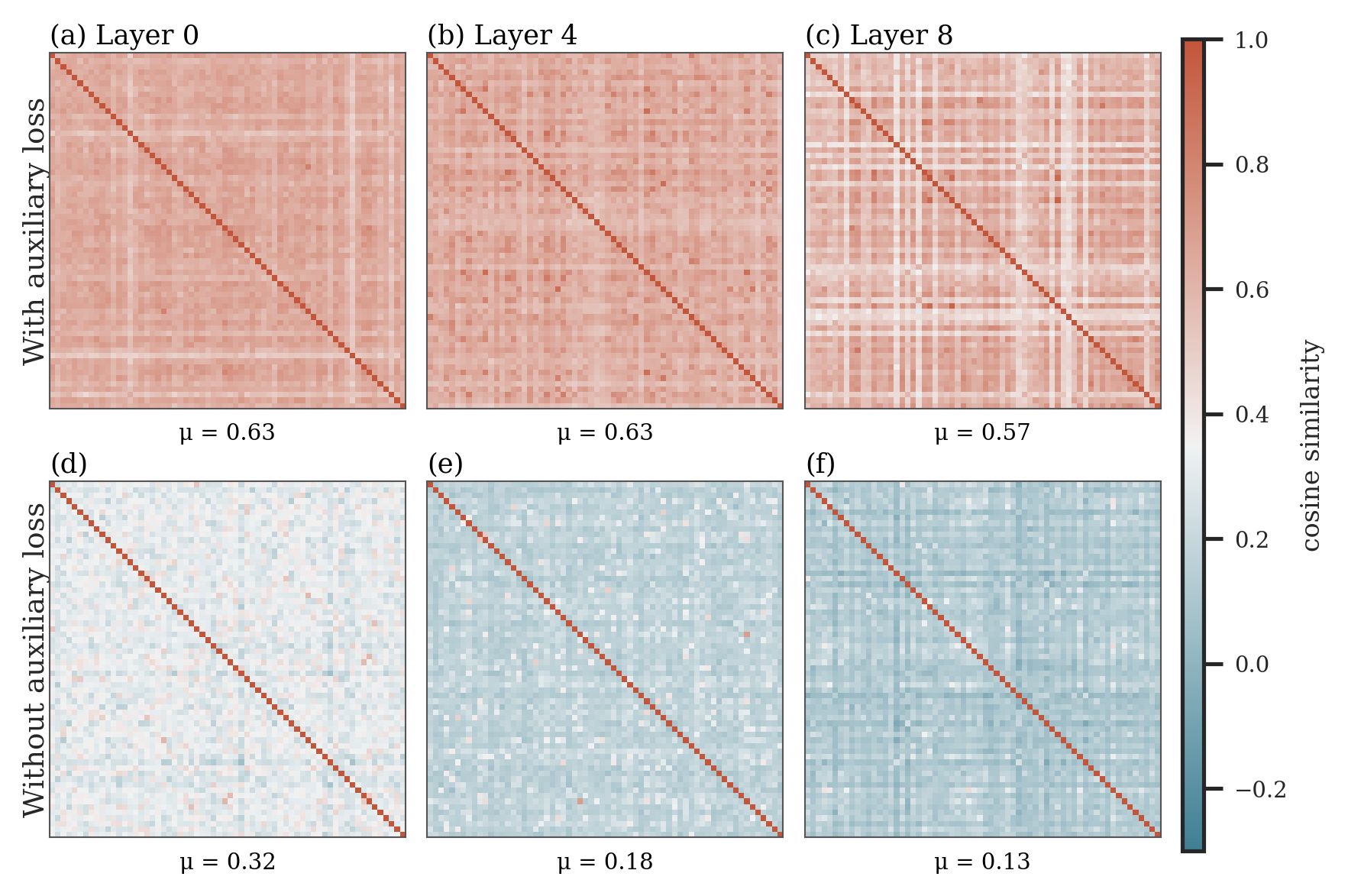}
  \caption{\textbf{Auxiliary loss collapses router geometry.}
  Each panel shows pairwise cosine similarities between router weight vectors within one model layer. The top row uses auxiliary load-balancing loss. The bottom row uses bias-only balancing with the same $1$B architecture and training setup. Off-diagonal means ($\mu$) appear below each panel.}
  \label{fig:router_heatmaps}
\end{figure}

\paragraph{Experimental setup}

We use the $1$B SMoE configuration from \citet{wang2024auxiliarylossfreeloadbalancing}, and train it from scratch on the OLMoE-mix-0924~\citep{muennighoff2025olmoeopenmixtureofexpertslanguage} dataset for $\sim$$50$B tokens.
Each model has $L{=}9$ SMoE layers, hidden size $d{=}1024$, $N{=}64$ routed experts, top-$K{=}6$ routing, $N_s{=}2$ shared experts, and expert hidden width $512$.
Using this configuration, we train two $1$B SMoEs from scratch, starting from the same initialization and training for the same number of tokens. 
The models use the same SMoE architecture and differ only in the balancing rule.
The run \textbf{without auxiliary loss} uses the bias-only balancing rule~\citep{wang2024auxiliarylossfreeloadbalancing}, where no balancing gradients reach the router.
The run \textbf{with auxiliary loss} replaces the bias rule with the auxiliary load-balancing loss used by Switch Transformers~\citep{fedus2022switchtransformersscalingtrillion} and the router $z$-loss from ST-MoE~\citep{zoph2022stmoedesigningstabletransferable}, a recipe used in recent SMoEs such as OLMoE and Mixtral~\citep{muennighoff2025olmoeopenmixtureofexpertslanguage,jiang2024mixtralexperts}.
After training, we extract the router weight matrix $W_r\in\mathbb{R}^{N\times d}$ at three stratified layers and report pairwise cosine similarities between its router weight vectors.

\paragraph{The auxiliary loss collapses router weight vectors}
Figure~\ref{fig:router_heatmaps} presents the results, showing a collapse in router geometry. With the auxiliary loss, the off-diagonal mean cosine similarity is $\mu = 0.63$, $0.63$, and $0.57$ at layers $0$, $4$, and $8$, respectively. By contrast, under loss-free bias balancing the same architecture remains more diverse, with substantially lower similarity scores of $\mu = 0.32$, $0.18$, and $0.13$. 
Overall, this showcases that auxiliary load-balancing aligns otherwise distinct router weight vectors, making them nearly three times more similar.

Section~\ref{sec:empirical_coupling} showed that router scores carry information when coupling is intact.
Here we identify a mechanism that erodes this information at the router itself.
When router weight vectors become strongly aligned, the expert-specific differences between router scores shrink relative to their shared component.
The softmax therefore has less expert-specific signal to use for distinguishing between experts.
Prior work has shown that interference gradients from auxiliary loss can impair language modeling \citep{wang2024auxiliarylossfreeloadbalancing} and degrade expert specialization~\citep{guo2025advancingexpertspecialization}.
Our measurement provides a geometric explanation for this degradation, localizing the effect directly in the router's own weight vectors.
This motivates the design we develop next, a router whose directions come directly from the inputs each expert has processed, without balancing gradients.

\section{Centroid tracking captures a substantial component of learned router--expert coupling}
\label{sec:kmeans_router}

The geometric coupling suggests that router weight vectors accumulate toward summaries of the inputs routed to each expert.
This turns the gradient dynamics into a form of online centroid tracking, where each router direction is repeatedly pulled toward the hidden states assigned to its expert and gradually becomes a running summary of that expert's routed-token cluster.

If routing is largely centroid tracking, then much of the router's role should be recoverable without trainable routing weights or router-side gradients. 
We test this by replacing the standard gating weights with non-learnable centroids updated as exponential moving averages of the hidden states assigned to each expert.
The resulting router has no trainable routing parameters and it absorbs no balancing gradients. In what follows, we describe our router in detail, and evaluate its performance compared to existing routing approaches.
    
\paragraph{Centroid update rule}
Our router adapts online clustering to SMoE routing. At every layer, we maintain one centroid per routed expert, corresponding to the router weight vectors in standard SMoEs.
Here, as before, the total number of centroids and experts is $N$, and the number of experts selected per token is $k$. 
For a token with hidden state $\mathbf{x}$, expert $i$ receives the score
\begin{equation}
s_i(\mathbf{x}) = \frac{\mathbf{c}_i^\top \mathbf{x}}{\|\mathbf{c}_i\|\,\|\mathbf{x}\|} + b_i,
\end{equation}
where $\mathbf{c}_i\in\mathbb{R}^d$ is the router's centroid corresponding to expert $i$ and $b_i$ is a scalar bias.
The token is routed to the $k$ experts with the largest scores.
Centroids are updated at every step by an exponential moving average over the inputs assigned to each expert,
\begin{equation}
\mathbf{c}_i \leftarrow \alpha\,\mathbf{c}_i + (1-\alpha)\,\bar{\mathbf{x}}_i,\qquad \bar{\mathbf{x}}_i = \tfrac{1}{|\mathcal{T}_i|}\textstyle\sum_{\mathbf{x}\in\mathcal{T}_i}\mathbf{x},
\label{eq:centroid_update}
\end{equation}
where $\mathcal{T}_i$ is the set of tokens routed to expert $i$ in the current micro-batch.

We update the biases using the loss-free rule of \citet{wang2024auxiliarylossfreeloadbalancing}, $b_i\leftarrow b_i + \gamma\,\mathrm{sign}(\tau - f_i)$, where $f_i$ is the realized load fraction, $\tau=k/N$ is the target load, and $\gamma > 0$ is the update rate.
Both centroids and biases are gradient-free running statistics, rather than trainable parameters.
Thus, the router makes the predicted centroid-tracking dynamics explicit, with no trainable router weights, no auxiliary loss, and no gradient flow into the routing decision.

\paragraph{Experimental setup}
To evaluate our new router, we use the same $1$B SMoE configuration as in Section~\ref{sec:aux_loss_collapse},
and train it from scratch on OLMoE-mix-0924 \citep{muennighoff2025olmoeopenmixtureofexpertslanguage} multiple times with different routing rules. Each run trains for $21$k steps ($50$B tokens), using AdamW with $\beta_1{=}0.9$, $\beta_2{=}0.95$, a peak learning rate of $10^{-3}$ decayed to $10^{-4}$, a $1$k-step warmup, and a batch size of $2.36$M tokens per step.
We compare four routing variants:
\begin{itemize}
    \item \textbf{Aux-Loss}: Uses the Switch-style auxiliary load-balancing loss together with the router $z$-loss regularization~\citep{fedus2022switchtransformersscalingtrillion,zoph2022stmoedesigningstabletransferable}. This is the same model analyzed in Section~\ref{sec:aux_loss_collapse}.
    \item \textbf{Loss-Free}: Uses the bias-only balancing rule of \citet{wang2024auxiliarylossfreeloadbalancing}, adjusting per-expert routing biases while leaving router weights free of balancing gradients.
    \item \textbf{Loss-Free + Seq-Aux}: In addition to the bias-based balancing of Loss-Free, it employs the sequence-wise auxiliary loss of DeepSeek-V3 \citep{deepseekai2025deepseekv3technicalreport}, which encourages each sequence to distribute its routed tokens more evenly across experts. This discourages within-sequence routing bottlenecks.
    \item \textbf{K-Means}: Our router, which follows the natural geometric coupling, using the centroid rule above with $\alpha{=}0.99$ and $\gamma{=}10^{-3}$. These hyperparameters were chosen so the centroid and bias updates are similar in magnitude to the other three variants.
\end{itemize}
We evaluate these variants in terms of language modeling performance and load imbalance. For language modeling, we report perplexity (PPL) over the C4-en and the Pile held-out validation slices from OLMo, derived from the C4~\citep{raffel2023exploringlimitstransferlearning} and Pile corpora~\citep{gao2020pile800gbdatasetdiverse}. For load-balancing, we use the MaxVio metric by \citet{wang2024auxiliarylossfreeloadbalancing}, defined as $\mathrm{MaxVio}=\max_i f_i/\bar f - 1$ and averaged across the nine layers. MaxVio measures the worst relative overload: a value of $0$ means perfectly balanced routing, while larger values indicate that at least one expert receives more tokens than the average expert.

\begin{table}[t]
\centering
\caption{\textbf{K-Means achieves comparable perplexity to the loss-free family and yields the lowest steady-state load imbalance.} All four runs use the $1$B SMoE configuration from \citet{wang2024auxiliarylossfreeloadbalancing} and are trained for $50$B tokens; they differ only in the routing rule. ``Router params'' counts the learnable parameters of the router, summed over the nine layers; the centroids and biases used by K-Means are non-learnable running statistics. MaxVio is averaged across the nine layers at step $21$k.}
\label{tab:kmeans_main}
\setlength{\tabcolsep}{3pt}
\begin{tabularx}{\columnwidth}{@{}>{\raggedright\arraybackslash}p{92pt} >{\centering\arraybackslash}p{47pt} >{\centering\arraybackslash}p{49pt} *{4}{>{\centering\arraybackslash}X}@{}}
\toprule
Method & Router parameters & Auxiliary losses  & Train PPL & C4-en PPL & Pile PPL & MaxVio\\
\midrule
Aux-Loss              & $0.59$M & load balance $+\,z$ & $15.09$ & $20.54$ & $11.82$ & $0.526$ \\
Loss-Free + Seq-Aux   & $0.59$M & seq-aux   & $15.03$ & $20.44$ & $11.77$ & $0.102$ \\
Loss-Free             & $0.59$M & --      & $15.01$ & $20.40$ & $11.76$ & $0.084$ \\
K-Means (ours)        & $0$     & --      & $15.40$ & $21.01$ & $12.09$ & $\mathbf{0.037}$ \\
\bottomrule
\end{tabularx}
\end{table}

\begin{figure}[t]
  \centering
  \includegraphics[width=\columnwidth]{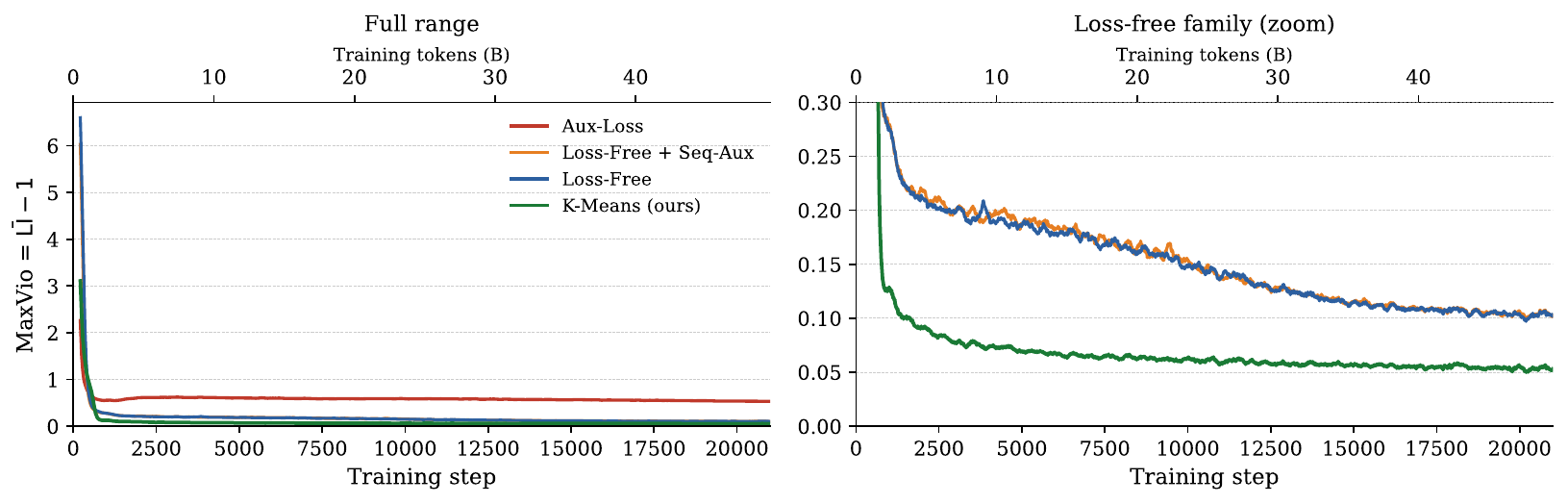}
  \caption{\textbf{Load imbalance during training (log scale).} Layer-averaged $\mathrm{MaxVio}$ versus training step for the four routing variants on the $1$B SMoE of \citet{wang2024auxiliarylossfreeloadbalancing}. Curves are $200$-step rolling means; the logarithmic $y$-axis separates the Aux-Loss collapse from the loss-free family while keeping both visible. K-Means settles to the lowest plateau ($\mathrm{MaxVio}\approx0.037$ at step $21$k) without any learned routing parameters or balancing gradient.}
  \label{fig:loadimb_curves}
\end{figure}

\paragraph{Results}

Table~\ref{tab:kmeans_main} reports perplexity and steady-state load imbalance of the trained models (using their last checkpoints).
K-Means achieves the lowest sustained load imbalance across variants, with $\mathrm{MaxVio}=0.037$ compared to $0.084$ for Loss-Free and $0.526$ for Aux-Loss, despite using no trainable routing parameters or balancing gradients.
This improvement comes with a modest perplexity cost relative to the learned loss-free router, with training PPL increasing by $2.6\%$ ($15.01 \rightarrow 15.40$) and similar gaps on C4-en and Pile.
Figure~\ref{fig:loadimb_curves} further shows the $\mathrm{MaxVio}$ score over training. The loss-free variants rapidly reduce imbalance and remain stable, while K-Means settles to the lowest and smoothest plateau.
Since its routing state consists only of exponential moving average centroids and load-balancing biases, the bias rule regulates expert usage without altering centroid directions through gradients.

\paragraph{Discussion}
Sections~\ref{sec:co_evolution_hyp}--\ref{sec:aux_loss_collapse} suggest that router weight vectors accumulate toward summaries of the inputs processed by their corresponding experts, while auxiliary balancing losses perturb this expert-specific accumulation.
Under this view, a router that explicitly tracks these centroids, without trainable routing weights or balancing gradients, should train stably and balance at least as well as the bias-only baseline. Our results support this prediction, as K-Means achieves tighter balance than Loss-Free with a $2.6\%$ training perplexity cost. 
We interpret the perplexity gap as the cost of removing routing degrees of freedom beyond centroid tracking, and its modest size as evidence that centroid tracking captures a substantial component of what gradient-trained routers learn.

\paragraph{Where the perplexity gap comes from}
Since K-Means centroids are means of each expert's routed hidden states, they should represent typical tokens well but not rare tokens or tokens lying between expert clusters.
A per-token comparison of negative log-likelihood (NLL) against Loss-Free at step $21$k on the full held-out suite ($6.9$M tokens) supports both predictions.
The NLL gap is $+0.061$ on the rarest $10\%$ of target tokens and vanishes on the most frequent ones, and it is $2.7\times$ larger when routing is consistently borderline across layers ($+0.044$ vs.\ $+0.016$ in the narrowest vs.\ widest decile of the layer-averaged margin between the $6$th and $7$th ranked scores).
This cost is specific to the centroid router: binned by their own margins, the learned routers show a narrowest-to-widest difference of only $+0.005$ versus $+0.027$ for K-Means, and the gap between the two learned routers, binned by the K-Means margins, is flat.

\section{Related work}
 
\paragraph{Geometric analyses of SMoE routing and experts}
Recent work analyzes SMoEs along several dimensions, including expert specialization, knowledge attribution, and routing behavior \citep{guo2025advancingexpertspecialization,li-etal-2025-decoding,yang2025moex,herbst2026expertstrikes}. 
Closest to ours are geometric analyses of expert and routing structure. \citet{liu2023diversifyingmoe} promote expert diversity through an orthogonal optimizer, and \citet{lo-etal-2025-closer} observe correlated router and expert gate-projection weights in pretrained models. Concurrent with our work, \citet{huang2026sdmoe} document aligned expert subspaces from shared low-rank input structure.
We complement these empirical analyses with a gradient-level account. \citet{dikkala-etal-2023-benefits-route} study router--expert correspondence under clustered-data assumptions, and \citet{lv2025coupling} enforce it through an explicit alignment loss; we show that this correspondence instead emerges under standard SMoE training from shared input-directed gradients between matched router rows and expert input weights, and is visible in expert activations.

\paragraph{Decoupled and constrained routing}
Many routing methods intervene on the router separately from expert weight evolution, either through auxiliary load-balancing losses \citep{shazeer2017outrageouslylargeneuralnetworks,lepikhin2020gshardscalinggiantmodels,fedus2022switchtransformersscalingtrillion} or bias-based balancing \citep{wang2024auxiliarylossfreeloadbalancing,deepseekai2025deepseekv3technicalreport}. 
Other works decouple routing more directly: \citet{dai2022stablemoestableroutingstrategy} freeze a distilled router during later training, \citet{sukhbaatar2024branchtrainmixmixingexpertllms} train experts independently before merging them into an MoE, \citet{pan2024densetrainingsparseinference} defer sparse routing to inference, and \citet{xu2026grouterdecouplingroutingrepresentation} distill fixed routing structures from pretrained models.
Our analysis suggests that such interventions should be evaluated not only by load balance, but also by whether they preserve router--expert geometry: auxiliary balancing can inject interference gradients into router directions, while stronger decoupling may discard the shared input-directed updates through which router rows and expert weights co-evolve.

\paragraph{Geometric and centroid-based routing}
Recent works replace learned MoE routing with explicit geometric structure, using shared eigenbases, per-expert subspaces, Grassmannian structure, or hidden-state subspaces to route tokens \citep{cheng2026emoe,cheng2025ermoe,shihab2026grassmannian,mohamud2026selfrouting}. 
Most similar to our K-Means router, \citet{yang2025lpr} route by cosine similarity to EMA-updated prototypes, though theirs are learned in a regularized latent projection space rather than computed as non-learnable averages of routed hidden states.
Centroid-based routing has also appeared in different settings, including online spherical $k$-means for sparse attention \citep{roy2021routingtransformer} and EMA centroid routing for frozen-backbone adapters \citep{prottasha2026monkeyjump}, neither targeting SMoE feed-forward routing during language-model pretraining.
These works impose geometric structure on routing through architectural design. We show, in contrast, that such structure already emerges in standard learned SMoE routers, since matched router vectors and expert input weights co-accumulate the hidden-state directions assigned to their experts.
To test this mechanism directly, our K-Means router replaces learned router weights with non-learnable EMA centroids of routed hidden states.

\section{Conclusion and discussion}

We show that learned routing in SMoEs is not independent of expert evolution but emerges from gradient dynamics shared with the experts it selects. Routers and the experts they select co-evolve as coupled accumulators of the tokens routed to each expert. This theory-driven geometric coupling also appears empirically in the model's computation, is fragile under interventions that bypass it, and largely recoverable from the routed token stream alone. Together, our findings suggest that router–expert coupling is a substantial part of what gradient-trained routers learn, and that SMoE training should preserve this geometry rather than perturb it.

\paragraph{Limitations and future work}

Our comparison of routing methods in Section~\ref{sec:kmeans_router} is limited to a single 1B SMoE configuration. Whether the observed perplexity and load-balancing trade-offs persist at larger scales and in other architectures remains open.
Beyond scale, our analysis of how auxiliary load-balancing gradients reduce separation between router directions captures only one aspect of router--expert geometry; SMoE training involves other pathologies and routing interventions, including representation collapse, expert dominance, and alternative balancing mechanisms.
A broader direction is to test whether the same geometric lens can help explain these phenomena by measuring how they preserve, distort, or bypass router--expert coupling.
Last, the K-Means router is a constructive test of the coupling view rather than a practical router; its perplexity gap suggests that learned routers use degrees of freedom beyond centroid position, motivating hybrid designs that preserve the centroid geometry while closing this gap.

\section*{Acknowledgments}
We thank Elad Shikley for his help with figure design, and Or Shafran for feedback on the manuscript.

{
    \small
    \bibliographystyle{unsrtnat}
    \bibliography{custom}
}

\appendix

\section{Theoretical foundation: the geometric alignment principle}
\label{sec:appendix_derivation}

To better understand the meaning of the router and expert weights, we analyzed how these weights are updated during training, focusing on the gradients involved.

\subsection{Setup}
\label{sec:appendix_setup}

\begin{table}[h]
    \caption{Mathematical notations and setup parameters.}
    \label{tab:appendix-setup}
    \centering
    \renewcommand{\arraystretch}{1.5}
    \begin{tabularx}{\textwidth}{>{\centering\arraybackslash}p{0.12\textwidth} >{\centering\arraybackslash}p{0.12\textwidth} X >{\centering\arraybackslash}p{0.12\textwidth}}
        \toprule
        \textbf{Category} & \textbf{Notation} & \textbf{Description} & \textbf{Dimension} \\
        \midrule
        \textbf{Input} & $\mathbf{x}$ & The input token representation (or hidden state). & $\mathbb{R}^d$ \\
        \midrule
        \textbf{Router} & $W_r$ & The router weight matrix. & $\mathbb{R}^{N \times d}$ \\
        & $\mathbf{r}_i$ & The $i$-th row of $W_r$; the router weight vector for expert $i$. & $\mathbb{R}^d$ \\
        & $z_i$ & The routing score (logit) for the $i$-th expert: \newline $z_i = x^\top \mathbf{r}_i$ & $\mathbb{R}$ \\
        & $p_i$ & The gating probability assigned to expert $i$, obtained via softmax over the logits: \newline $p_i = e^{z_i} / \sum_j e^{z_j}$ & $\mathbb{R}$ \\
        \midrule
        \textbf{Expert} & $W^{\mathrm{gate}}_{i}$ & The gate projection matrix for expert $i$. & $\mathbb{R}^{d_{ff} \times d}$ \\
        & $W^{\mathrm{up}}_{i}$ & The up-projection matrix for expert $i$. & $\mathbb{R}^{d_{ff} \times d}$ \\
        & $W^{\mathrm{down}}_{i}$ & The down-projection matrix for expert $i$. & $\mathbb{R}^{d \times d_{ff}}$ \\
        & $E_i(x)$ & The output of the $i$-th expert (SwiGLU): \newline $E_i(x) = W^{\mathrm{down}}_{i}(\mathrm{SiLU}(W^{\mathrm{gate}}_{i}\,x) \odot W^{\mathrm{up}}_{i}\,x)$ & $\mathbb{R}^d$ \\
        \midrule
        \textbf{Output} & $y$ & The final SMoE layer output, calculated as the weighted sum of the expert outputs: \newline $y = \sum_j p_j E_j(x)$ & $\mathbb{R}^d$ \\
        \midrule
        \textbf{Loss} & $\mathcal{L}_y := \frac{\partial \mathcal{L}}{\partial y}$ & The upstream gradient, representing the derivative of the total loss $\mathcal{L}$ with respect to the output $y$. & $\mathbb{R}^d$ \\
        \bottomrule
    \end{tabularx}
\end{table}

\subsection{Expert gradient derivation}

We derive the gradient for $W^{\mathrm{up}}_{i}$. In SwiGLU, the expert computes
\begin{equation}
E_i(x) = W^{\mathrm{down}}_{i}\,h_i, \qquad h_i = g_i \odot W^{\mathrm{up}}_{i}\,x, \qquad g_i = \mathrm{SiLU}(W^{\mathrm{gate}}_{i}\,x).
\end{equation}
Crucially, $W^{\mathrm{up}}_{i}\,x$ enters the computation \emph{linearly}: the gate branch $g_i$ depends on $W^{\mathrm{gate}}_{i}$, not on $W^{\mathrm{up}}_{i}$. This is precisely why we analyze $W^{\mathrm{up}}_{i}$---the gradient path through the linear branch avoids activation-function derivatives entirely, yielding a clean input-directed update.

The SMoE output is $y = \sum_j p_j E_j(x)$, so $\frac{\partial y}{\partial E_i} = p_i$. Applying the chain rule:
\begin{enumerate}
    \item \textbf{Through the SMoE weighted sum and down-projection:}
    \begin{equation}
    \frac{\partial \mathcal{L}}{\partial h_i}
    = (W^{\mathrm{down}}_{i})^T\!\left(p_i\,\mathcal{L}_y\right) \in \mathbb{R}^{d_{ff}}
    \end{equation}

    \item \textbf{Through the Hadamard gate (linear in $W^{\mathrm{up}}_{i}\,x$):}
    Since $h_i = g_i \odot W^{\mathrm{up}}_{i}\,x$ and $g_i$ does not depend on $W^{\mathrm{up}}_{i}$,
    \begin{equation}
    \frac{\partial \mathcal{L}}{\partial (W^{\mathrm{up}}_{i}\,x)}
    = \frac{\partial \mathcal{L}}{\partial h_i} \odot g_i \in \mathbb{R}^{d_{ff}}
    \end{equation}

    \item \textbf{Final matrix gradient (outer product with input):}
    \begin{equation}
    \frac{\partial \mathcal{L}}{\partial W^{\mathrm{up}}_{i}}
    = \frac{\partial \mathcal{L}}{\partial (W^{\mathrm{up}}_{i}\,x)}\; x^T \in \mathbb{R}^{d_{ff} \times d}
    \end{equation}
\end{enumerate}

The $k$-th row of $W^{\mathrm{up}}_{i}$ (hidden neuron $u_k$) therefore receives the gradient
\begin{equation}
\frac{\partial \mathcal{L}}{\partial u_k} = \delta_{i,k}\; x^T \in \mathbb{R}^{1 \times d},
\qquad
\delta_{i,k} = \underbrace{\left[(W^{\mathrm{down}}_{i})^T\!\left(p_i\,\mathcal{L}_y\right)\right]_k}_{\text{error signal}} \cdot\; \underbrace{g_{i,k}}_{\text{gate value}}
\end{equation}
where $\delta_{i,k}$ is the scalar error term for neuron $k$, which absorbs the upstream loss gradient, the routing probability, the down-projection, and the gate activation---all factors that are independent of $W^{\mathrm{up}}_{i}$. The directional component of every row update is $x^T$: each row of $W^{\mathrm{up}}_{i}$ evolves as a weighted sum of the input representations it processes.

\subsection{Router gradient derivation}
\label{sec:router_appendix_derivation}

The routing logit for expert $i$ is $z_i = x^T \mathbf{r}_i$, where $\mathbf{r}_i$ is the $i$-th row of $W_r$. From the SMoE output $y = \sum_j p_j E_j(x)$, the loss gradient with respect to $p_i$ is:
\begin{equation}
\frac{\partial \mathcal{L}}{\partial p_i} = \mathcal{L}_y^T E_i(x) \in \mathbb{R}
\end{equation}

Since $\frac{\partial z_i}{\partial \mathbf{r}_i} = x$, the chain rule gives:
\begin{equation}
\frac{\partial \mathcal{L}}{\partial \mathbf{r}_i}
= \underbrace{\left(\mathcal{L}_y^T E_i(x)\right) \frac{\partial p_i}{\partial z_i}}_{\gamma_i}\; x
= \gamma_i\; x \in \mathbb{R}^{d}
\end{equation}
where $\gamma_i$ is a scalar that absorbs the upstream loss gradient, the expert output, and the softmax Jacobian---all independent of $\mathbf{r}_i$ itself.

\subsection{The shared structure}

Both gradients have the same form: a scalar error signal multiplied by the input $\mathbf{x}$.
\begin{align}
\frac{\partial \mathcal{L}}{\partial u_k} &= \delta_{i,k}\; x^T && \text{(expert row)} \\[4pt]
\frac{\partial \mathcal{L}}{\partial \mathbf{r}_i} &= \gamma_i\; x && \text{(router weight vector)}
\end{align}
The scalars $\delta_{i,k}$ and $\gamma_i$ differ---they encode different error signals---but the directional component is the same input $\mathbf{x}$ in both cases. Over training, each row of $W^{\mathrm{up}}_{i}$ and each router weight vector $\mathbf{r}_i$ accumulates a weighted sum of the hidden states routed to expert $i$. Because matched router--expert pairs process the same token stream, this shared input-directed update structure predicts that they will develop statistically aligned geometry in hidden space.

\subsection{Extension to AdamW}
\label{sec:adamw_derivation}

For the task loss, the raw gradients retain the forms $\gamma_i x$ for $\mathbf{r}_i$ and $\delta_{i,k}x^T$ for $u_k$ under AdamW. Its first moments accumulate weighted histories of these inputs, while coordinate-wise second-moment rescaling changes the update directions. Momentum and weight decay can also update parameters with zero current gradients. Shared input histories therefore motivate geometric coupling under AdamW without guaranteeing aligned router and expert weights.

\section{Resources}
\label{sec:resources}
All experiments were run on an Nvidia H100 node, or an AMD MI325X node.

\section{Preactivation measurements}
\label{app:preactivation}
We provide further details on the measurements in Figure~\ref{fig:expert_router_correlation}. All three models are evaluated on the same 1,000 English Wikipedia article titles. We curated the titles with AI assistance, selecting 200 from each of five domains: science and technology, nature and health, history and geography, arts and notable people, and society and everyday life. We used Wikipedia metadata to resolve canonical titles and exclude missing, duplicate and disambiguation pages.

For each title, we measure the final token. For every selected expert at each MoE layer, we record the router score and the $L_2$ norms of its gate and up preactivations. We standardize scores and norms separately within each layer and expert, then pool the selected pairs into eight router-score quantile bins. Each curve shows the median norm at the mean score in each bin, with the interquartile range shaded.

Both larger models show an overall increase in median gate and up preactivation norms with router score (Figure~\ref{fig:expert_router_correlation}). These measurements concern selected token--expert pairs.

\subsection{Selected and non-selected experts}
\label{app:selected-nonselected}

We compare the gate and up preactivation norms of selected and non-selected experts on the same token representations. For each token and layer, we compute $\|W_i^{\mathrm{gate}}\mathbf{x}\|_2$ and $\|W_i^{\mathrm{up}}\mathbf{x}\|_2$ for every expert in the routed-expert pool, including those not selected, without multiplying by routing weights. For each layer, expert, and projection, we normalize the norms using the mean and population standard deviation over all valid tokens, shared by the selected and non-selected groups. We then average within each group and over tokens, and equally over layers. Selection follows each model's native router. This analysis uses all valid tokens from the 1,000 article-title inputs: 3,303 tokens for OLMoE, 3,151 for Qwen, and 3,020 for DeepSeek, rather than only the final tokens used in Figure~\ref{fig:expert_router_correlation}.

Table                                           ~\ref{tab:selected-nonselected} reports the selected-minus-non-selected difference. OLMoE has positive differences for both projections in every layer. Qwen has a positive gate difference in every layer, but a slightly negative average up difference. DeepSeek has a positive average up difference, while its gate difference is small and varies across layers. Thus, higher preactivation norms among selected experts depend on the model, projection, and layer.

\begin{table}[htbp]
\centering
\caption{Selected-minus-non-selected mean preactivation norms after shared layer--expert normalization. Differences are in standardized units and averaged equally over layers. Layer counts are descriptive, not significance tests.}
\label{tab:selected-nonselected}
\begin{tabular}{lrrrr}
\toprule
& \multicolumn{2}{c}{Mean difference} & \multicolumn{2}{c}{Layers with positive difference} \\
Model & Gate & Up & Gate & Up \\
\midrule
OLMoE & $+0.471$ & $+0.292$ & 16/16 & 16/16 \\
Qwen & $+0.166$ & $-0.014$ & 48/48 & 13/48 \\
DeepSeek & $+0.008$ & $+0.281$ & 22/40 & 39/40 \\
\bottomrule
\end{tabular}

\end{table}

\section{Training dynamics}
\label{app:dynamics}
We analyze eleven publicly released OLMoE training checkpoints, spanning steps 5k--1.22M. Since the earliest checkpoint we analyze is after 5,000 steps, we also follow our loss-free model from initialization.

\subsection{OLMoE}
We use the same 1,000 titles at all eleven checkpoints and compute Spearman correlations between router scores and gate and up preactivation norms, standardized within each layer and expert. The gate correlation increases up to 20,000 steps, whereas the up correlation is already close to its highest measured value at 5,000 steps. Both decline after 20,000 steps while remaining positive.

\begin{figure}[t]
\centering
\includegraphics[width=.9\linewidth]{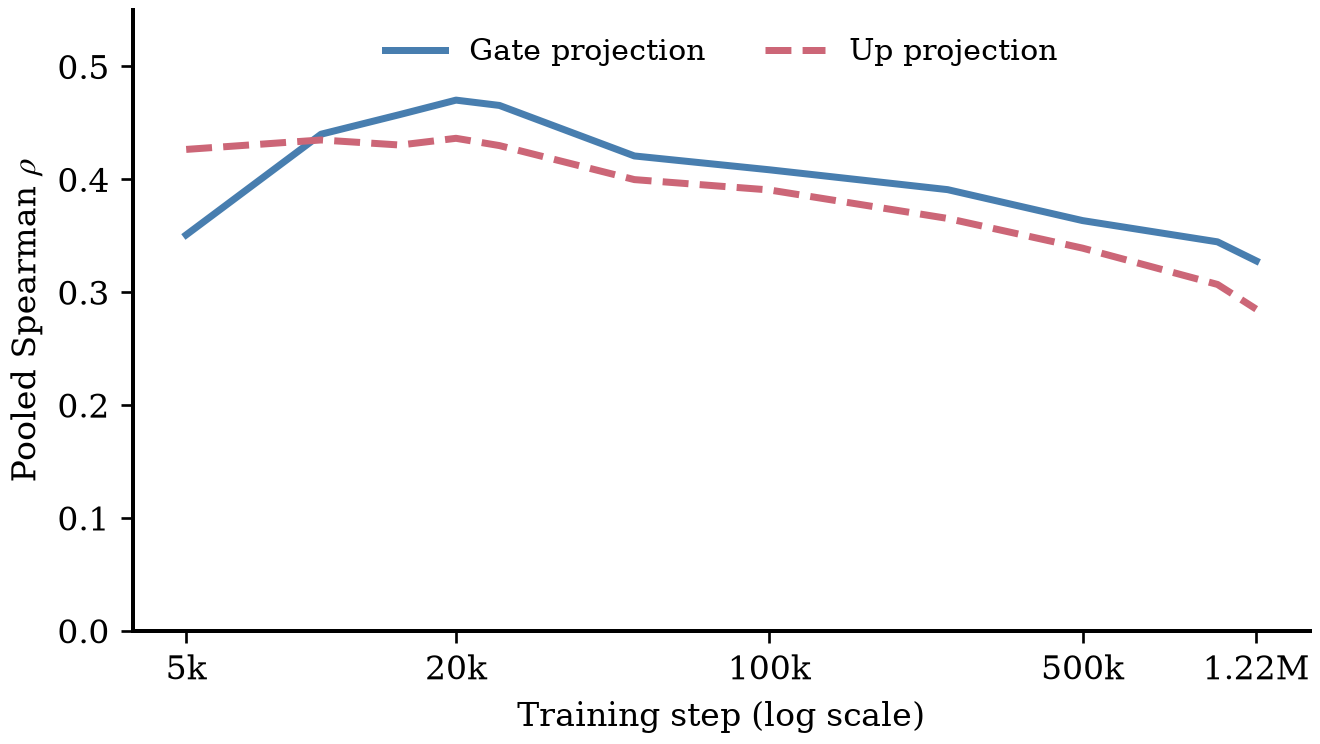}
\caption{Router--expert correlation during OLMoE training. Spearman correlation between router scores and gate and up preactivation $L_2$ norms for the same 1,000 title-final tokens at each checkpoint. Scores and norms are standardized within layer and expert over selected pairs. The training-step axis is logarithmic.}
\label{fig:olmoe_dynamics}
\end{figure}

\subsection{Our loss-free model}
We follow six checkpoints from initialization to 40K steps, corresponding to about 93.8 billion training tokens. For each checkpoint, we measure mean post-SiLU gate activation at the final token of 1,024 titles. For routed pairs, the correlation rises early and then declines, remaining above the non-routed and permuted-pairing controls after initialization.

To test whether the correlation depends on matching each router vector to its expert, we randomly permute expert identities within each layer and recompute the correlation.

\begin{figure}[t]
\centering
\includegraphics[width=.85\linewidth]{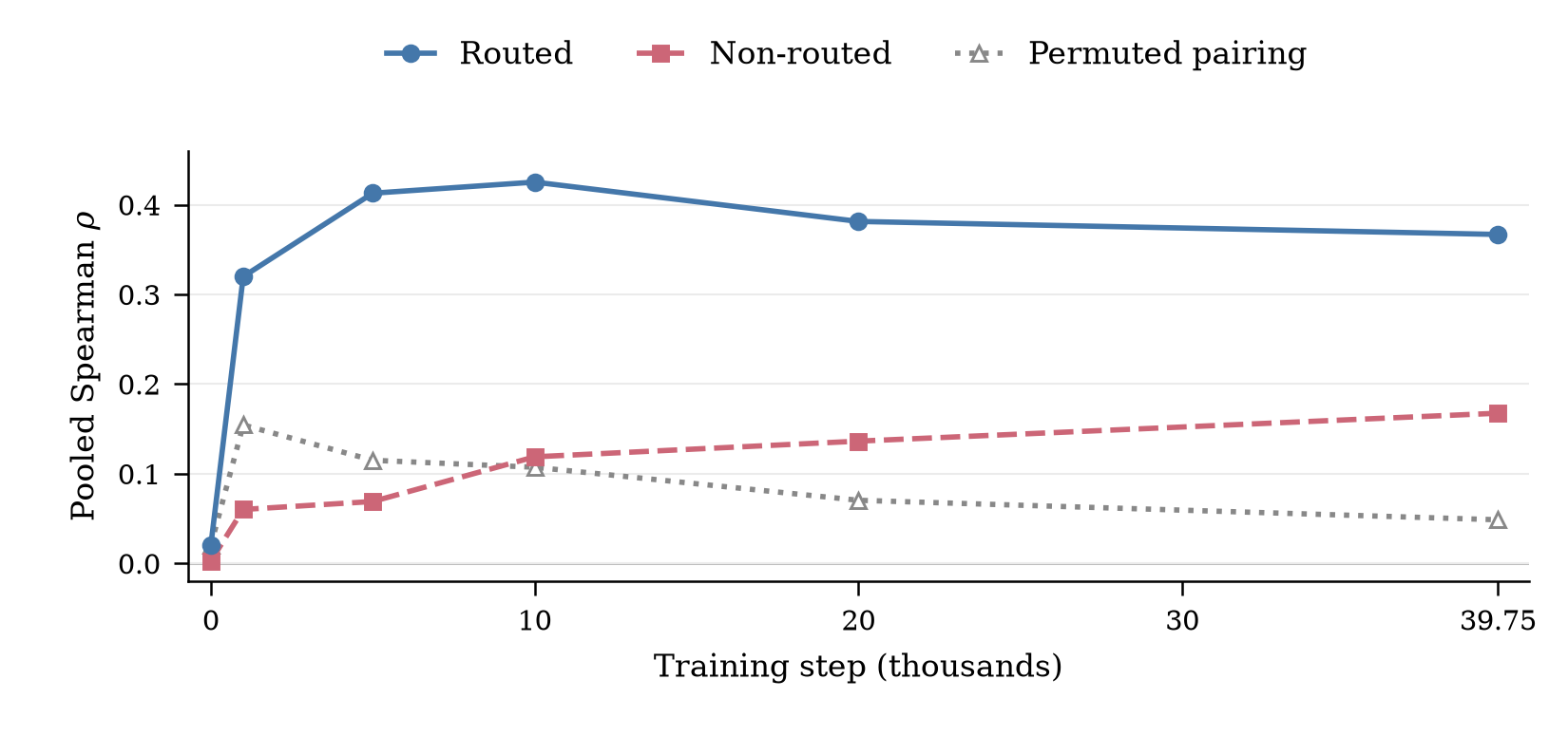}
\caption{Router--expert correlation during training of our loss-free model. Spearman correlation between router scores and mean gate-neuron preactivations for 1,024 title-final tokens. Routed and non-routed pairs are standardized separately within layer and expert. The permuted-pairing control averages correlations from twenty expert-identity permutations.}
\label{fig:own_dynamics}
\end{figure}

\section{Removing router-aligned input components}
\label{app:causal}
We intervene at one OLMoE layer at a time, after routing. Router and control directions are normalized to unit length, and each intervention removes the full projection of the expert input along its direction. We apply the intervention to all selected experts at that layer, keeping its expert selection and routing weights fixed; subsequent layers run normally.

For each expert, we construct 32 control directions orthogonal to its router vector using the gate and up weight matrices. We evaluate every direction on the same 176 validation sequences, with 512 next-token predictions per sequence and bfloat16 inference. We report the change in mean negative log-likelihood from the unmodified model.

\begin{table}[t]
\caption{Change in mean negative log-likelihood after removing input components in OLMoE, relative to the unmodified model ($\Delta$NLL, nats/token). We report the router-direction intervention and the median across 32 perpendicular controls at each layer. Router rank orders all 33 interventions by decreasing loss increase (1 = largest). Positive values indicate worse predictions.}
\label{tab:causal}
\centering
\setlength{\tabcolsep}{10pt}
\begin{tabular}{rrrr}
\toprule
Layer & Router $\Delta$NLL & Median control $\Delta$NLL & Rank / 33 \\
\midrule
1 & 0.00097 & 0.00048 & 11 \\
2 & 0.00376 & 0.00060 & 3 \\
3 & 0.02412 & 0.00054 & 2 \\
4 & 0.02217 & 0.00046 & 2 \\
5 & 0.00546 & 0.00039 & 2 \\
6 & 0.00200 & 0.00037 & 3 \\
7 & 0.00287 & 0.00055 & 2 \\
8 & 0.00276 & 0.00024 & 2 \\
9 & 0.01541 & 0.00050 & 4 \\
10 & 0.00953 & 0.00053 & 3 \\
11 & 0.03890 & 0.00137 & 2 \\
12 & 0.08915 & 0.00164 & 2 \\
13 & -0.00171 & 0.00142 & 33 \\
14 & 0.12756 & 0.00566 & 1 \\
15 & 0.17778 & 0.11184 & 5 \\
16 & 0.01384 & 0.00285 & 5 \\
\bottomrule
\end{tabular}
\end{table}

\clearpage
\section*{NeurIPS Paper Checklist}
\begin{enumerate}

\item {\bf Claims}
    \item[] Question: Do the main claims made in the abstract and introduction accurately reflect the paper's contributions and scope?
    \item[] Answer: \answerYes{}
    \item[] Justification: The abstract and introduction summarize the paper's main theoretical claim of router--expert geometric coupling, the empirical evidence in Section~\ref{sec:empirical_coupling} and Appendix~\ref{app:dynamics}, the auxiliary-loss effect analysis, and the parameter-free centroid-router experiment.
    \item[] Guidelines:
    \begin{itemize}
        \item The answer \answerNA{} means that the abstract and introduction do not include the claims made in the paper.
        \item The abstract and/or introduction should clearly state the claims made, including the contributions made in the paper and important assumptions and limitations. A \answerNo{} or \answerNA{} answer to this question will not be perceived well by the reviewers. 
        \item The claims made should match theoretical and experimental results, and reflect how much the results can be expected to generalize to other settings. 
        \item It is fine to include aspirational goals as motivation as long as it is clear that these goals are not attained by the paper. 
    \end{itemize}

\item {\bf Limitations}
    \item[] Question: Does the paper discuss the limitations of the work performed by the authors?
    \item[] Answer: \answerYes{}
    \item[] Justification: The paper includes a ``Limitations and future work'' paragraph discussing the routing-method comparison on a single $1$B SMoE configuration, the need to validate across larger scales and architectures, and the remaining small perplexity gap of the K-Means router.
    \item[] Guidelines:
    \begin{itemize}
        \item The answer \answerNA{} means that the paper has no limitation while the answer \answerNo{} means that the paper has limitations, but those are not discussed in the paper. 
        \item The authors are encouraged to create a separate ``Limitations'' section in their paper.
        \item The paper should point out any strong assumptions and how robust the results are to violations of these assumptions (e.g., independence assumptions, noiseless settings, model well-specification, asymptotic approximations only holding locally). The authors should reflect on how these assumptions might be violated in practice and what the implications would be.
        \item The authors should reflect on the scope of the claims made, e.g., if the approach was only tested on a few datasets or with a few runs. In general, empirical results often depend on implicit assumptions, which should be articulated.
        \item The authors should reflect on the factors that influence the performance of the approach. For example, a facial recognition algorithm may perform poorly when image resolution is low or images are taken in low lighting. Or a speech-to-text system might not be used reliably to provide closed captions for online lectures because it fails to handle technical jargon.
        \item The authors should discuss the computational efficiency of the proposed algorithms and how they scale with dataset size.
        \item If applicable, the authors should discuss possible limitations of their approach to address problems of privacy and fairness.
        \item While the authors might fear that complete honesty about limitations might be used by reviewers as grounds for rejection, a worse outcome might be that reviewers discover limitations that aren't acknowledged in the paper. The authors should use their best judgment and recognize that individual actions in favor of transparency play an important role in developing norms that preserve the integrity of the community. Reviewers will be specifically instructed to not penalize honesty concerning limitations.
    \end{itemize}

\item {\bf Theory assumptions and proofs}
    \item[] Question: For each theoretical result, does the paper provide the full set of assumptions and a complete (and correct) proof?
    \item[] Answer: \answerYes{}
    \item[] Justification: The paper states the SMoE architecture and training setup used for the gradient analysis in Section~\ref{sec:smoe_arch_and_training}. The main text gives the gradient-level argument for router--expert coupling, and Appendix~\ref{sec:appendix_derivation} provides the full derivations for the expert and router gradients.
    \item[] Guidelines:
    \begin{itemize}
        \item The answer \answerNA{} means that the paper does not include theoretical results. 
        \item All the theorems, formulas, and proofs in the paper should be numbered and cross-referenced.
        \item All assumptions should be clearly stated or referenced in the statement of any theorems.
        \item The proofs can either appear in the main paper or the supplemental material, but if they appear in the supplemental material, the authors are encouraged to provide a short proof sketch to provide intuition. 
        \item Inversely, any informal proof provided in the core of the paper should be complemented by formal proofs provided in appendix or supplemental material.
        \item Theorems and Lemmas that the proof relies upon should be properly referenced. 
    \end{itemize}

    \item {\bf Experimental result reproducibility}
    \item[] Question: Does the paper fully disclose all the information needed to reproduce the main experimental results of the paper to the extent that it affects the main claims and/or conclusions of the paper (regardless of whether the code and data are provided or not)?
    \item[] Answer: \answerYes{}
    \item[] Justification: The paper describes the $1$B SMoE architecture, training setup, datasets, optimization hyperparameters, routing variants, balancing rules, and evaluation metrics used for the main experiments. The router-score versus expert-activation analysis, auxiliary-loss comparison, and centroid-router update rule are specified in Sections~\ref{sec:empirical_coupling}, \ref{sec:aux_loss_collapse}, and~\ref{sec:kmeans_router}.
    \item[] Guidelines:
    \begin{itemize}
        \item The answer \answerNA{} means that the paper does not include experiments.
        \item If the paper includes experiments, a \answerNo{} answer to this question will not be perceived well by the reviewers: Making the paper reproducible is important, regardless of whether the code and data are provided or not.
        \item If the contribution is a dataset and\slash or model, the authors should describe the steps taken to make their results reproducible or verifiable. 
        \item Depending on the contribution, reproducibility can be accomplished in various ways. For example, if the contribution is a novel architecture, describing the architecture fully might suffice, or if the contribution is a specific model and empirical evaluation, it may be necessary to either make it possible for others to replicate the model with the same dataset, or provide access to the model. In general. releasing code and data is often one good way to accomplish this, but reproducibility can also be provided via detailed instructions for how to replicate the results, access to a hosted model (e.g., in the case of a large language model), releasing of a model checkpoint, or other means that are appropriate to the research performed.
        \item While NeurIPS does not require releasing code, the conference does require all submissions to provide some reasonable avenue for reproducibility, which may depend on the nature of the contribution. For example
        \begin{enumerate}
            \item If the contribution is primarily a new algorithm, the paper should make it clear how to reproduce that algorithm.
            \item If the contribution is primarily a new model architecture, the paper should describe the architecture clearly and fully.
            \item If the contribution is a new model (e.g., a large language model), then there should either be a way to access this model for reproducing the results or a way to reproduce the model (e.g., with an open-source dataset or instructions for how to construct the dataset).
            \item We recognize that reproducibility may be tricky in some cases, in which case authors are welcome to describe the particular way they provide for reproducibility. In the case of closed-source models, it may be that access to the model is limited in some way (e.g., to registered users), but it should be possible for other researchers to have some path to reproducing or verifying the results.
        \end{enumerate}
    \end{itemize}

\item {\bf Open access to data and code}
    \item[] Question: Does the paper provide open access to the data and code, with sufficient instructions to faithfully reproduce the main experimental results, as described in supplemental material?
    \item[] Answer: \answerYes{}
    \item[] Justification: We provide code in the supplemental material, including the implementation of the router-score versus expert-activation analysis, the auxiliary-loss comparison, and the centroid-router update rule. The paper and supplemental material describe the model configuration, training setup, datasets, routing rules, and evaluation procedures needed to reproduce the main experimental results.
    \item[] Guidelines:
    \begin{itemize}
        \item The answer \answerNA{} means that paper does not include experiments requiring code.
        \item Please see the NeurIPS code and data submission guidelines (\url{https://neurips.cc/public/guides/CodeSubmissionPolicy}) for more details.
        \item While we encourage the release of code and data, we understand that this might not be possible, so \answerNo{} is an acceptable answer. Papers cannot be rejected simply for not including code, unless this is central to the contribution (e.g., for a new open-source benchmark).
        \item The instructions should contain the exact command and environment needed to run to reproduce the results. See the NeurIPS code and data submission guidelines (\url{https://neurips.cc/public/guides/CodeSubmissionPolicy}) for more details.
        \item The authors should provide instructions on data access and preparation, including how to access the raw data, preprocessed data, intermediate data, and generated data, etc.
        \item The authors should provide scripts to reproduce all experimental results for the new proposed method and baselines. If only a subset of experiments are reproducible, they should state which ones are omitted from the script and why.
        \item At submission time, to preserve anonymity, the authors should release anonymized versions (if applicable).
        \item Providing as much information as possible in supplemental material (appended to the paper) is recommended, but including URLs to data and code is permitted.
    \end{itemize}

\item {\bf Experimental setting/details}
    \item[] Question: Does the paper specify all the training and test details (e.g., data splits, hyperparameters, how they were chosen, type of optimizer) necessary to understand the results?
    \item[] Answer: \answerYes{}
    \item[] Justification: The paper specifies the model configuration, training data, number of training tokens and steps, optimizer, learning-rate schedule, warmup, batch size, routing variants, balancing rules, and evaluation metrics in Sections~\ref{sec:empirical_coupling}, \ref{sec:aux_loss_collapse}, and~\ref{sec:kmeans_router}. Additional implementation details are provided in the supplemental code.
    \item[] Guidelines:
    \begin{itemize}
        \item The answer \answerNA{} means that the paper does not include experiments.
        \item The experimental setting should be presented in the core of the paper to a level of detail that is necessary to appreciate the results and make sense of them.
        \item The full details can be provided either with the code, in appendix, or as supplemental material.
    \end{itemize}

\item {\bf Experiment statistical significance}
    \item[] Question: Does the paper report error bars suitably and correctly defined or other appropriate information about the statistical significance of the experiments?
    \item[] Answer: \answerYes{}
    \item[] Justification: Figure~\ref{fig:expert_router_correlation} reports median responses with shaded interquartile ranges, showing variability across token--expert observations within each router-score bin. Appendix~\ref{app:preactivation} describes the within-layer/expert standardization, quantile binning, and calculation of these summaries.
    \item[] Guidelines:
    \begin{itemize}
        \item The answer \answerNA{} means that the paper does not include experiments.
        \item The authors should answer \answerYes{} if the results are accompanied by error bars, confidence intervals, or statistical significance tests, at least for the experiments that support the main claims of the paper.
        \item The factors of variability that the error bars are capturing should be clearly stated (for example, train/test split, initialization, random drawing of some parameter, or overall run with given experimental conditions).
        \item The method for calculating the error bars should be explained (closed form formula, call to a library function, bootstrap, etc.)
        \item The assumptions made should be given (e.g., Normally distributed errors).
        \item It should be clear whether the error bar is the standard deviation or the standard error of the mean.
        \item It is OK to report 1-sigma error bars, but one should state it. The authors should preferably report a 2-sigma error bar than state that they have a 96\% CI, if the hypothesis of Normality of errors is not verified.
        \item For asymmetric distributions, the authors should be careful not to show in tables or figures symmetric error bars that would yield results that are out of range (e.g., negative error rates).
        \item If error bars are reported in tables or plots, the authors should explain in the text how they were calculated and reference the corresponding figures or tables in the text.
    \end{itemize}

\item {\bf Experiments compute resources}
    \item[] Question: For each experiment, does the paper provide sufficient information on the computer resources (type of compute workers, memory, time of execution) needed to reproduce the experiments?
    \item[] Answer: \answerYes{}
    \item[] Justification: We detail the resources in \ref{sec:resources}. For our experiments we used AMD MI325X accelerators. We used a single AMD MI325X GPU per experiment. The total tokens throughput for the mentioned configuration was 10B tokens a day, and each run lasted 5 days on a single card. All our experiments were contained and constrained to the compute mentioned above.
    \item[] Guidelines:
    \begin{itemize}
        \item The answer \answerNA{} means that the paper does not include experiments.
        \item The paper should indicate the type of compute workers CPU or GPU, internal cluster, or cloud provider, including relevant memory and storage.
        \item The paper should provide the amount of compute required for each of the individual experimental runs as well as estimate the total compute. 
        \item The paper should disclose whether the full research project required more compute than the experiments reported in the paper (e.g., preliminary or failed experiments that didn't make it into the paper). 
    \end{itemize}
    
\item {\bf Code of ethics}
    \item[] Question: Does the research conducted in the paper conform, in every respect, with the NeurIPS Code of Ethics \url{https://neurips.cc/public/EthicsGuidelines}?
    \item[] Answer: \answerYes{}
    \item[] Justification: We have reviewed the NeurIPS Code of Ethics and believe the research conforms to it. The work studies model training dynamics and routing mechanisms, uses standard language-modeling datasets and evaluations, and does not involve human subjects or sensitive personal data collection.
    \item[] Guidelines:
    \begin{itemize}
        \item The answer \answerNA{} means that the authors have not reviewed the NeurIPS Code of Ethics.
        \item If the authors answer \answerNo, they should explain the special circumstances that require a deviation from the Code of Ethics.
        \item The authors should make sure to preserve anonymity (e.g., if there is a special consideration due to laws or regulations in their jurisdiction).
    \end{itemize}

\item {\bf Broader impacts}
    \item[] Question: Does the paper discuss both potential positive societal impacts and negative societal impacts of the work performed?
    \item[] Answer: \answerNo{}
    \item[] Justification: The work is foundational and not tied to a specific deployment. Any societal impacts are indirect through broader improvements to language-model training.
    \item[] Guidelines:
    \begin{itemize}
        \item The answer \answerNA{} means that there is no societal impact of the work performed.
        \item If the authors answer \answerNA{} or \answerNo, they should explain why their work has no societal impact or why the paper does not address societal impact.
        \item Examples of negative societal impacts include potential malicious or unintended uses (e.g., disinformation, generating fake profiles, surveillance), fairness considerations (e.g., deployment of technologies that could make decisions that unfairly impact specific groups), privacy considerations, and security considerations.
        \item The conference expects that many papers will be foundational research and not tied to particular applications, let alone deployments. However, if there is a direct path to any negative applications, the authors should point it out. For example, it is legitimate to point out that an improvement in the quality of generative models could be used to generate Deepfakes for disinformation. On the other hand, it is not needed to point out that a generic algorithm for optimizing neural networks could enable people to train models that generate Deepfakes faster.
        \item The authors should consider possible harms that could arise when the technology is being used as intended and functioning correctly, harms that could arise when the technology is being used as intended but gives incorrect results, and harms following from (intentional or unintentional) misuse of the technology.
        \item If there are negative societal impacts, the authors could also discuss possible mitigation strategies (e.g., gated release of models, providing defenses in addition to attacks, mechanisms for monitoring misuse, mechanisms to monitor how a system learns from feedback over time, improving the efficiency and accessibility of ML).
    \end{itemize}
    
\item {\bf Safeguards}
    \item[] Question: Does the paper describe safeguards that have been put in place for responsible release of data or models that have a high risk for misuse (e.g., pre-trained language models, image generators, or scraped datasets)?
    \item[] Answer: \answerNA{}
    \item[] Justification: The paper does not release a high-risk model or new scraped dataset. The released code is for reproducing routing analyses and experiments.
    \item[] Guidelines:
    \begin{itemize}
        \item The answer \answerNA{} means that the paper poses no such risks.
        \item Released models that have a high risk for misuse or dual-use should be released with necessary safeguards to allow for controlled use of the model, for example by requiring that users adhere to usage guidelines or restrictions to access the model or implementing safety filters. 
        \item Datasets that have been scraped from the Internet could pose safety risks. The authors should describe how they avoided releasing unsafe images.
        \item We recognize that providing effective safeguards is challenging, and many papers do not require this, but we encourage authors to take this into account and make a best faith effort.
    \end{itemize}

\item {\bf Licenses for existing assets}
    \item[] Question: Are the creators or original owners of assets (e.g., code, data, models), used in the paper, properly credited and are the license and terms of use explicitly mentioned and properly respected?
    \item[] Answer: \answerYes{}
    \item[] Justification: The paper cites the existing models, datasets, and methods used in the experiments in Sections~\ref{sec:empirical_coupling}, \ref{sec:aux_loss_collapse}, and~\ref{sec:kmeans_router}. We respect the terms of use of these assets and include license information for released code and external assets in the supplemental material.
    \item[] Guidelines:
    \begin{itemize}
        \item The answer \answerNA{} means that the paper does not use existing assets.
        \item The authors should cite the original paper that produced the code package or dataset.
        \item The authors should state which version of the asset is used and, if possible, include a URL.
        \item The name of the license (e.g., CC-BY 4.0) should be included for each asset.
        \item For scraped data from a particular source (e.g., website), the copyright and terms of service of that source should be provided.
        \item If assets are released, the license, copyright information, and terms of use in the package should be provided. For popular datasets, \url{paperswithcode.com/datasets} has curated licenses for some datasets. Their licensing guide can help determine the license of a dataset.
        \item For existing datasets that are re-packaged, both the original license and the license of the derived asset (if it has changed) should be provided.
        \item If this information is not available online, the authors are encouraged to reach out to the asset's creators.
    \end{itemize}

\item {\bf New assets}
    \item[] Question: Are new assets introduced in the paper well documented and is the documentation provided alongside the assets?
    \item[] Answer: \answerYes{}
    \item[] Justification: We release code as a new asset in the supplemental material, with documentation for reproducing the routing analyses and experiments. The released asset does not include new datasets or human-subject data.
    \item[] Guidelines:
    \begin{itemize}
        \item The answer \answerNA{} means that the paper does not release new assets.
        \item Researchers should communicate the details of the dataset\slash code\slash model as part of their submissions via structured templates. This includes details about training, license, limitations, etc. 
        \item The paper should discuss whether and how consent was obtained from people whose asset is used.
        \item At submission time, remember to anonymize your assets (if applicable). You can either create an anonymized URL or include an anonymized zip file.
    \end{itemize}

\item {\bf Crowdsourcing and research with human subjects}
    \item[] Question: For crowdsourcing experiments and research with human subjects, does the paper include the full text of instructions given to participants and screenshots, if applicable, as well as details about compensation (if any)? 
    \item[] Answer: \answerNA{}
    \item[] Justification: The paper does not involve crowdsourcing experiments or research with human subjects.
    \item[] Guidelines:
    \begin{itemize}
        \item The answer \answerNA{} means that the paper does not involve crowdsourcing nor research with human subjects.
        \item Including this information in the supplemental material is fine, but if the main contribution of the paper involves human subjects, then as much detail as possible should be included in the main paper. 
        \item According to the NeurIPS Code of Ethics, workers involved in data collection, curation, or other labor should be paid at least the minimum wage in the country of the data collector. 
    \end{itemize}

\item {\bf Institutional review board (IRB) approvals or equivalent for research with human subjects}
    \item[] Question: Does the paper describe potential risks incurred by study participants, whether such risks were disclosed to the subjects, and whether Institutional Review Board (IRB) approvals (or an equivalent approval/review based on the requirements of your country or institution) were obtained?
    \item[] Answer: \answerNA{}
    \item[] Justification: The paper does not involve crowdsourcing experiments or research with human subjects, so IRB approval is not applicable.
    \item[] Guidelines:
    \begin{itemize}
        \item The answer \answerNA{} means that the paper does not involve crowdsourcing nor research with human subjects.
        \item Depending on the country in which research is conducted, IRB approval (or equivalent) may be required for any human subjects research. If you obtained IRB approval, you should clearly state this in the paper. 
        \item We recognize that the procedures for this may vary significantly between institutions and locations, and we expect authors to adhere to the NeurIPS Code of Ethics and the guidelines for their institution. 
        \item For initial submissions, do not include any information that would break anonymity (if applicable), such as the institution conducting the review.
    \end{itemize}

\item {\bf Declaration of LLM usage}
    \item[] Question: Does the paper describe the usage of LLMs if it is an important, original, or non-standard component of the core methods in this research? Note that if the LLM is used only for writing, editing, or formatting purposes and does \emph{not} impact the core methodology, scientific rigor, or originality of the research, declaration is not required.
    \item[] Answer: \answerYes{}
    \item[] Justification: Appendix~\ref{app:preactivation} describes AI-assisted curation of the 1,000 Wikipedia titles used as evaluation inputs, including the domain allocation and subsequent checks against Wikipedia metadata. This use contributes to evaluation-input selection and is distinct from writing, editing or formatting assistance.
    \item[] Guidelines:
    \begin{itemize}
        \item The answer \answerNA{} means that the core method development in this research does not involve LLMs as any important, original, or non-standard components.
        \item Please refer to our LLM policy in the NeurIPS handbook for what should or should not be described.
    \end{itemize}

\end{enumerate}

\end{document}